\documentclass[final]{hooml2022} 

\usepackage{graphicx}
\usepackage{amsmath,amssymb}
\usepackage{array}
\usepackage{color}
\usepackage{nowidow}
\usepackage[ruled]{algorithm2e}
\definecolor{mydarkblue}{rgb}{0,0.08,0.45}
\newcommand{\x}{\mathbf{x}}

\usepackage{algorithm}
\usepackage{algpseudocode}

\usepackage{graphicx}
\usepackage{booktabs} 
\usepackage{bm}
\usepackage{stmaryrd}
\usepackage{multirow}
\newcommand{\STAB}[1]{\begin{tabular}{@{}c@{}}#1\end{tabular}}

\newtheorem{thm:thm}{Theorem}[section]
\newtheorem{thm:def}{Definition}[section]
\newtheorem{thm:lemma}{Lemma}[section]

\title[High-Order Optimization of Gradient Boosted Decision Trees]{High-Order Optimization of Gradient Boosted Decision Trees}

\optauthor{%
\Name{Jean Pachebat} \Email{jpachebat@gmail.com}\\
\Name{Sergey Ivanov} \Email{ivanovserg990@gmail.com}\\
\addr Criteo AI Lab, Paris, France
}

\begin{document}

\maketitle

\begin{abstract}%
Gradient Boosted Decision Trees (GBDTs) are dominant machine learning algorithms for modeling discrete or tabular data. Unlike neural networks with millions of trainable parameters, GBDTs optimize loss function in an additive manner and have a single trainable parameter per leaf, which makes it easy to apply high-order optimization of the loss function. In this paper, we introduce high-order optimization for GBDTs based on numerical optimization theory which allows us to construct trees based on high-order derivatives of a given loss function. In the experiments, we show that high-order optimization has faster per-iteration convergence that leads to reduced running time. Our solution can be easily parallelized and run on GPUs with little overhead on the code. Finally, we discuss future potential improvements such as automatic differentiation of arbitrary loss function and combination of GBDTs with neural networks. 
\end{abstract}


\section{Introduction}
Gradient boosted decision trees (GBDT)~\cite{friedman2000additive} are state-of-the-art ML models for tabular datasets. Many variants of GBDTs have been proposed in the recent years that achieve top performance in classification \cite{he2014practical, richardson2007predicting}, regression \cite{ivanov2021boost, chen2022does}, and ranking~\cite{pmlr-v14-gulin11a, ustimenko2019learning} tasks with applications where data often contain missing or noisy features, complex relationships, and heterogeneity such as in recommender systems, information retrieval, and many others \cite{chen2021task, qin2021are, islam2020evidence, ozcaglar2019entity, zhang2015gradient}.

Let’s consider a dataset $\mathcal{D} = \left\{ \left( {{\x_i},{y_i}} \right)|i \in \left\{ {1,...,n} \right\} \right\}$, where ${\x_i} \in {\mathbb{R}^m}$ and ${y_i} \in \mathbb{R}$. A GBDT model is a sum of $K$ additive functions, each of which represents a decision tree:
\begin{equation}\label{eq:tree}
\hat{y}_i = \phi(\x_i) = \sum^K_{k=1} f_k(\x_i), \ \ f_k\in \mathcal{F},
\end{equation}
where $\mathcal{F}=\{f(\x) = w_{q(\x)}\} ( q : \mathbb{R}^m \rightarrow T, w\in \mathbb{R}^T) $ is a space of regression trees. Each regression tree $f_k$ maps $m$-dimensional feature vector $\x$ to one of its leaves (or regions) with index $q(\x)$ that has a corresponding weight value $w_{q(\x)}$.

One of the most important differences between GBDT models and other tree-based models such as Random Forest \cite{breiman2001random} or Extremely Randomized Trees \cite{geurts2006extremely} is that GBDT model learns the trees in an iterative fashion by adding a new tree with respect to the performance of previous trees, while the latter models grow trees independently from each other. In particular, in GBDT the loss objective $\mathcal{L}^{(t)}$ at every iteration $t$ and regularization term $\Omega$ is minimized by adding a new tree $f_t$:

\begin{equation}\label{eq:loss}
\mathcal{L}^{(t)} = \sum_{i=1}^n l(y_i,\hat{y_i}^{(t-1)}+f_t(\x_i))+\Omega(f_t)
\end{equation}

The regularization term $\Omega$ often penalizes the complexity of the tree function and we consider it to be $\Omega(f_t) = \frac{1}{2} \lambda \|w\|^2$, where $\lambda$ is a hyperparameter. 

In the seminal paper~\cite{friedman2001greedy} Friedman proposed to learn new trees such that they produce the steepest-descent step direction given by the first-order gradient of the loss function with respect to the last prediction of the model: 
\begin{equation}\label{eq:first-order}
f_t = -\partial_{\hat{y}^{(t-1)}}l(y_i,\hat{y}^{(t-1)})
\end{equation}
Thus, each new tree gets an updated version of the data $\left\{ \left( {{\x_i},{\tilde{y}_i}} \right)|i \in \left\{ {1,...,n} \right\} \right\}$, where $\tilde{y}_i = \partial_{\hat{y}^{(t-1)}}l(y_i,\hat{y}^{(t-1)})$ is a “pseudoresponse” that the tree is fitting. 

More than a decade after Chen and Guestrin proposed XGBoost \cite{chen2016xgboost} that popularized second-order optimization of gradient boosting. Instead of fitting the first-order gradient, each new tree decomposes the loss function using second-order approximation and derives the closed-form formulas for assigning the leaf weights and building a regression tree. 

Let $g^{(k)}_i$ be the $k$-th order derivative of the loss function with respect to the last model's prediction  $\hat{y}^{(t-1)}$, i.e. $g^{(k)}_i = \partial^k_{\hat{y}^{(t-1)}}l(y_i,\hat{y}^{(t-1)})$. Then the Eq.~\ref{eq:loss} can be approximated by second-order Taylor expansion:

\begin{equation}\label{eq:second-order}
\mathcal{L}^{(t)} \simeq \sum_{i=1}^n [l(y_i,\hat{y}^{(t-1)}) + g^{(1)}_i f_t(\x_i)+\frac{1}{2}g^{(2)}_i f_t^2(\x_i)] + \Omega(f_t)
\end{equation}

Let’s first assume that the tree structure is given to us and we want to find the leaf weights that minimize the Eq.~\ref{eq:second-order}. Let $I_j=\{i|q(\x_i)=j\}$ be a set of all data instances that fall into the leaf $j$. Then, removing the term that does not depend on the tree $f_t$ and regrouping the data for each leaf we obtain: 
\begin{equation}\label{eq:regrouped}
\begin{split}
\tilde{\mathcal{L}}^{(t)}
         &=\sum^n_{i=1} [g^{(1)}_i f_t(\x_i)+\frac{1}{2}g^{(2)}_if_t^2(\x_i)] + \frac{1}{2}\lambda\sum^T_{j=1}w_j^2\\
         &=\sum^T_{j=1}[(\sum_{i\in I_j} g^{(1)}_i)w_j+\frac{1}{2}(\sum_{i\in I_j} g^{(2)}_i+\lambda)w_j^2]
\end{split}
\end{equation}

For each leaf $j$ let's denote $G^{(k)}_j = \sum_{i\in I_j} g^{(k)}_{i}$ be the sum of $k$-th order gradients of data instances that fall into that leaf. Then the minimum of Eq.~\ref{eq:regrouped} for each leaf $j$ is given by 
\begin{equation}\label{eq:leafscore}
w^*_j =-\frac{G^{(1)}_j}{G^{(2)}_j+\lambda},
\end{equation}
and by plugging it back to Eq.\ref{eq:regrouped} we get the optimal loss value: 
\begin{equation}\label{eq:optimalloss2}
\mathcal{L}^{*}_{j} = - \frac{1}{2} \sum^T_{j=1}\frac{[G^{(1)}_j]^2}{G^{(2)}_j + \lambda}.
\end{equation}


To date, the power of second-order GBDT models \cite{ustimenko2021sglb, sprangers2021probabilistic, prokhorenkova2018catboost, ke2017lightgbm, chen2016xgboost} has been demonstrated across a range of tasks and baselines including modern neural networks \cite{grinsztajn2022tree,borisov2021deep}. However, to the best of our knowledge, there are no works that study high-order gradient information of the loss function during optimization. In this work, we consider a high-order Taylor expansion of the loss function (Eq.~\ref{eq:loss}) and derive closed-form formulas for arbitrary order gradient statistics and compare the results of these higher-order methods in the experiments. 


\section{High-Order Optimization of GBDT}\label{part2}
We start by noting that XGboost decomposition of the loss function Eq.~\ref{eq:second-order} provides two advantages. First, it allows us to derive optimal leaf weights for arbitrary differentiable loss functions. Second, the closed-form Eq.~\ref{eq:leafscore} provides a greedy algorithm to construct a tree. However, the second-order Taylor approximation may lead to inaccurate estimation of the loss function and therefore longer convergence. Next, we show an example of the optimal leaf weights that use third-order approximation, which is followed by the general $k$-th order methods. 

\subsection{Cubic Optimization of GBDT}
Let's start with a third-order Taylor expansion of the loss Eq.~\ref{eq:loss} (after removing constant terms): 

\begin{equation}\label{eq:third-order}
\begin{split}
\tilde{\mathcal{L}}^{(t)}
         &=\sum^T_{j=1}[G^{(1)}_j w_j+\frac{1}{2}(G^{(2)}_j+\lambda)w_j^2 + \frac{1}{6}G^{(3)}_j w_j^3]
\end{split}
\end{equation}

Here $G^{(k)}_j = \sum_{i\in I_j} g^{(k)}_{i}$ be the sum of $k$-th order gradients of data instances in the leaf $j$. By equating the derivative to zero, we can find the optimal weights $w_j$ for each leaf $j$: 

\begin{equation}\label{eq:leafscore3}
w^*_j =-\frac{G^{(2)}_j+\lambda}{G^{(3)}_j}\left(1 \pm \sqrt{1 - \frac{2G^{(1)}_j G^{(3)}_j}{(G^{(2)}_j+\lambda)^2}}\right)
\end{equation}

Note that at the minimum the first-order gradients $G^{(1)}_j$ are zero and therefore for a small enough $\sum_{i\in I_j} g^{(1)}_i$  we can use an expansion $1 - \sqrt{1 - \alpha} = \frac{\alpha}{2} + \frac{\alpha^2}{8} + O(\alpha^3)$ to simplify the terms of this equation . The optimal weights become:

\begin{equation}\label{eq:leafscore3_approx}
w^*_j =-\frac{G^{(1)}_j}{G^{(2)}_j+\lambda}\left(1 + \frac{G^{(1)}_j G^{(3)}_j}{2(G^{(2)}_j+\lambda)^2}\right)
\end{equation}

By plugging the weights back into the Eq.~\ref{eq:third-order} we can compute the optimal loss value: 

\begin{equation}\label{eq:optimalloss3}
\begin{split}
\mathcal{L}^{*}_{j} =- \frac{[G^{(1)}_j]^2}{G^{(2)}_j + \lambda}(\frac{1}{2} + \frac{1}{6}\varepsilon + 2\varepsilon^3 + \frac{2}{3}\varepsilon^4)
 \end{split}
\end{equation}

Here, $\varepsilon = \dfrac{G^{(1)}_jG^{(3)}_j}{G^{(2)}_j+ \lambda}$. By comparing it to the Eq.~\ref{eq:optimalloss2} we can observe additional terms in Eq.~\ref{eq:optimalloss3} which correct the loss value estimation. Similarly, to XGBoost approach we can design an efficient greedy algorithm (Alg.~\ref{alg:exact-greedy}) that estimate the goodness of each split by computing the loss reduction score $\mathcal{L}_{split} = \mathcal{L}_{node} – (\mathcal{L}_{left}+\mathcal{L}_{right})$ and then selecting the split with the maximum score. To make it efficient, the algorithm first sorts the data on that node based on the feature values and then computes the loss based on the Eq.~\ref{eq:optimalloss3}. Similar algorithms can be designed for arbitrary order $k$. 

\subsection{Householder Optimization of GBDT}
Recap that the GBDT model is an additive process of adding new trees, each of which approximates some function of the loss objective $\phi(\x) \leftarrow \phi(\x) + f_t(\x)$. In the case of Friedman's first-order method $f_t(\x)$ approximates the gradient Eq.~\ref{eq:first-order}. In the second and third-order methods, each tree $f_t(\x)$ approximates functions that involve higher-order gradient statistics, given by Eqs.~\ref{eq:leafscore} and \ref{eq:leafscore3}. Furthermore, under some conditions of the regularity of the loss function, Householder~\cite{householder1970numerical} gave the general formula for arbitrary high order:

\begin{equation}
\label{eq:householder}
    \phi(\x) \leftarrow \phi(\x) + (p+1)\left(\frac{(1/g)^{(p)}}{(1/g)^{(p+1)}}\right)_{x_n},
\end{equation}

where $g$ is the gradient of the loss function and $(1/g)^{(p)}$ is the derivative of order $p$ of inverse of $g$. The convergence has an order $(p+2)$. For example, for $p=0$ Eq.~\ref{eq:householder} results in Newton-Raphson update of XGBoost (Eq.~\ref{eq:leafscore}). For $p=1$, Householder equation gives an update of order 3, also known as Halley's method~\cite{boyd2013finding}: 

\begin{equation}
\label{eq:halley}
    w^*_j =-\frac{G^{(1)}_j}{G^{(2)}_j+\lambda}\left(1 - \frac{G^{(1)}_j G^{(3)}_j}{2(G^{(2)}_j+\lambda)^2}\right)^{-1}
\end{equation}

Given the approximation $(1 - \alpha)^{-1} = 1 + \alpha + O(\alpha^2)$ for small $\alpha$, we can recover third-order update in Eq.~\ref{eq:leafscore3}. Finally, for $p=2$ we obtain the following fourth-order update: 

\begin{equation}
\label{eq:fourth-order}
\begin{split}
w^*_j =& -G^{(1)}_j\left(\frac{(G^{(2)}_j+\lambda)^2 - G^{(1)}_j G^{(3)}_j/2}{(G^{(2)}_j+\lambda)^3 -  G^{(1)}_j(G^{(2)}_j + \lambda)G^{(3)}_j + G^{(1)}_j G^{(4)}_j /6}\right).
 \end{split}
\end{equation}

\section{Experiments}

In the experiments, we use Eqs.~\ref{eq:leafscore}, \ref{eq:halley}, \ref{eq:fourth-order} for the second, third, and fourth-order models respectively. Our implementation is based on the open-source package Py-Boost~\footnote{https://github.com/jpachebat/py\_boost}. We used four binary classification datasets \cite{grinsztajn2022tree} for which we optimize binary cross-entropy loss. For each dataset, we use the validation part to select the best hyperparameters and use the test part to measure the accuracy.  To compare models' efficiency we first trained the second-order model (GBDT-2) for 10000 trees and then recorded the best test accuracy it achieves. Then for each model, we measured the running time it takes to reach 99\% of the best accuracy so that the small perturbations in the loss value can be discarded. The training was performed on GPU, with 120GB memory, 7.8Gbps, and 2 GPUs. We kept the maximum depth of 6 for trees. In contrary to second order, higher order GBDTs are more sensitive to regularization factor $\lambda$, which we selected using hyperparameter search in the range [$10^0, \ldots, 10^6$].

\begin{figure}[t]
\centering     
\subfigure[Epsilon]{\label{fig:epsilon}\includegraphics[width=.47\textwidth]{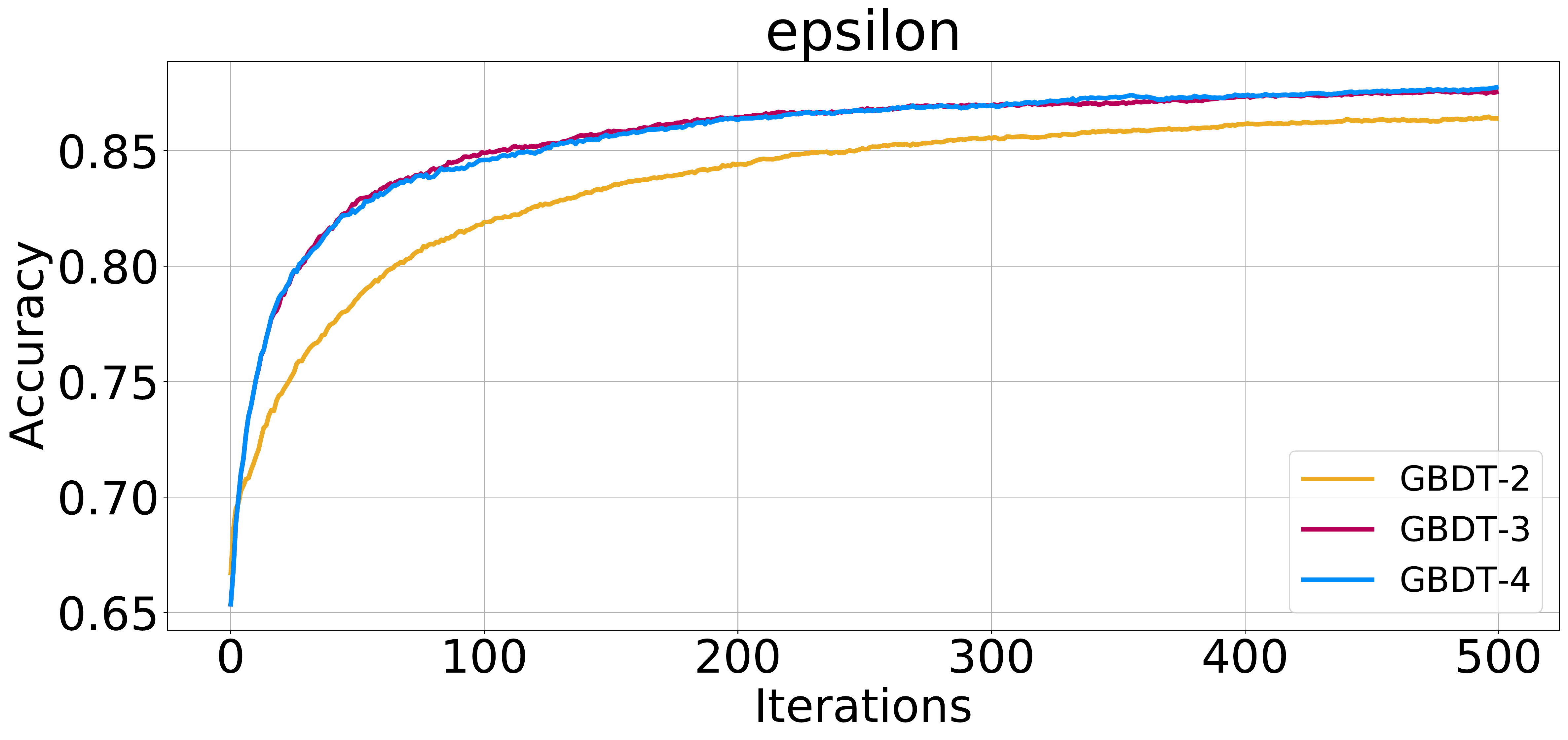}}
\subfigure[Higgs]{\label{fig:Higgs}\includegraphics[width=.47\textwidth]{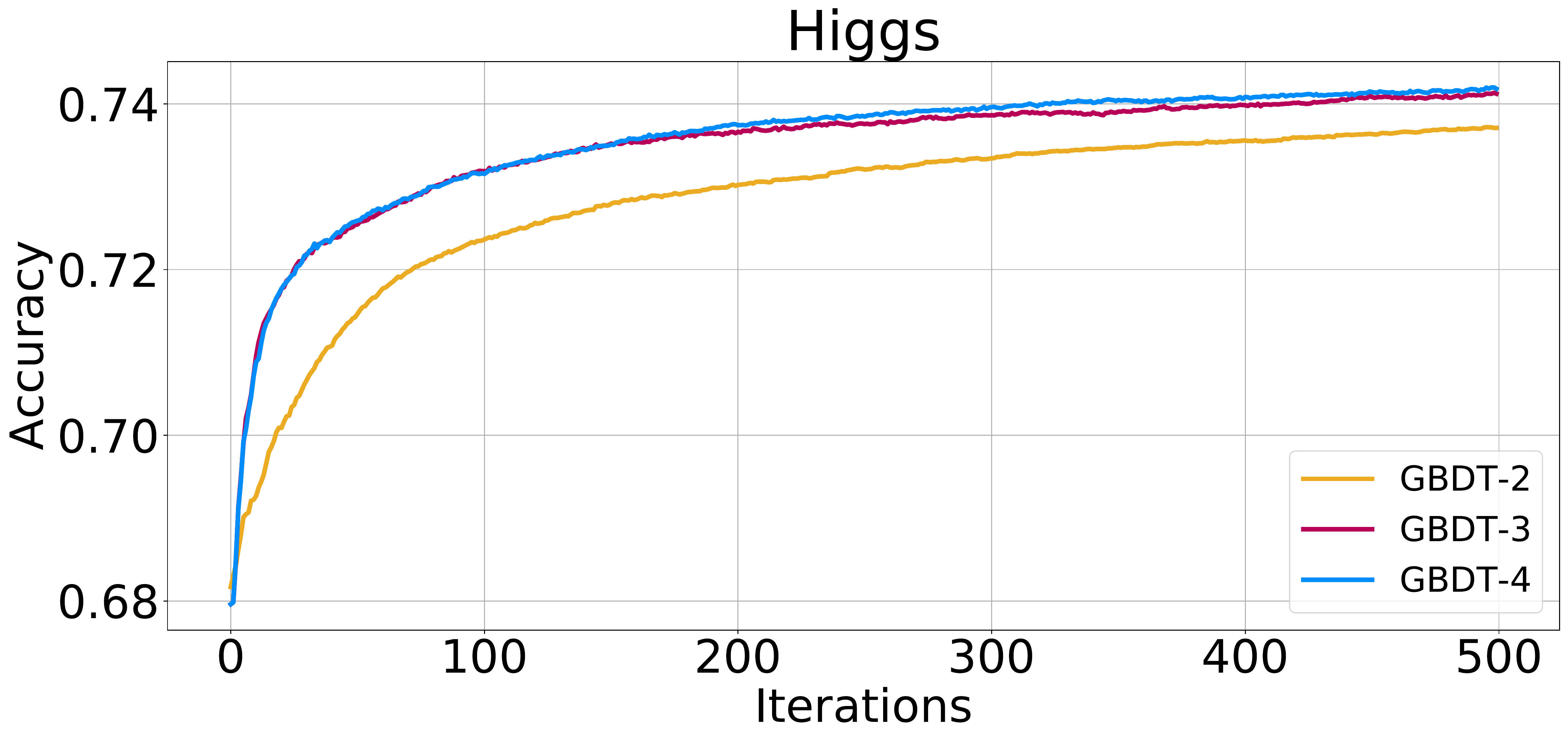}}
\caption{Test accuracy for high-order GBDT models.}
\label{fig:accuracy}
\end{figure}

Convergence of validation accuracy is presented in the Fig.~\ref{fig:accuracy}. We can see that the third (GBDT-3) and fourth (GBDT-4) methods have much faster convergence initially. However, as we train longer, the accuracy of GBDT-2 becomes comparable to those of the higher-order methods (see Appendix~\ref{app:longertraining}). On the other hand, the running time to reach 99\% optimal accuracy for higher-order is significantly smaller as shown in Table~\ref{tab:runtime}. For example, on Epsilon dataset the relative difference (Gap) between GBDT-2 model and GBDT-3 model is 73\%.

\begin{table*}[h]
\caption{Running time to achieve 99\% of the optimal test accuracy for high-order GBDT models. Gap \% is the relative difference w.r.t.~GBDT-2 accuracy (the smaller, the better).}\label{tab:runtime}
\vskip 0.15in
\begin{center}
\centering
\footnotesize
\resizebox{\textwidth}{!}{
\begin{tabular}{ll|rr|rr|rr|rr}
 & &  \multicolumn{2}{c|}{\textbf{Higgs}}  & \multicolumn{2}{c|}{\textbf{Epsilon}} & \multicolumn{2}{c|}{Covertype} & \multicolumn{2}{c}{MiniBooNE} \\
 &  Dataset &  
 \multicolumn{1}{c}{Time (s.)} & \multicolumn{1}{c|}{Gap \%}  & 
 \multicolumn{1}{c}{Time (s.)} & \multicolumn{1}{c|}{Gap \%} &
 \multicolumn{1}{c}{Time (s.)} & \multicolumn{1}{c|}{Gap \%} &
 \multicolumn{1}{c}{Time (s.)} & \multicolumn{1}{c}{Gap \%} \\
 \midrule
 \midrule
& GBDT-2        & 8.98 $\pm$ 0.01 & 0 & 94.80 $\pm$ 0.01 & 0 & 32.08 $\pm$ 0.01 & 0 & 1.35 $\pm$ 0.01 & 0  \\ \midrule
\multirow{2}{*}{\STAB{\rotatebox[origin=c]{90}{Ours}}}
& GBDT-3       & 4.93 $\pm$ 0.00 & -45 & 25.00 $\pm$ 0.01 & -73 & 18.46 $\pm$ 0.00 & -42 & 0.59 $\pm$ 0.01 & -56     \\
& GBDT-4       & 4.7 $\pm$ 0.1 & -47 & 29.17 $\pm$ 0.01 & -69 & 23.09 $\pm$ 0.00 & -28 & 0.48 $\pm$ 0.01 & -64   \\
 \bottomrule
\end{tabular}
}
\end{center}
\vskip -0.4in
\end{table*}

\section{Conclusion}
In this work, we proposed a high-order optimization framework to learn GBDT model, which has not been explored in the context of gradient boosting and may lead to many improvements to the existing GBDT algorithms: faster convergence, automatic regularization of the step size, and better optima. There are many exciting future directions for this research. Our framework is developed for arbitrary differentiable loss objectives; however, the user still has to provide manually-derived gradients in order to compute the optimal weights. Recent interest in automatic and symbolic differentiation \cite{baydin2018automatic} can come to the rescue, especially in the case when the loss objective is highly non-linear and the optimization order $p$ is high. Note, however, that automatic differentiation additionally increases the running time so there is a trade-off between the efficiency of higher-order optimization and its versatility. Going one step further, we can implement a GBDT model directly in popular frameworks such as PyTorch~\cite{paszke2019pytorch} or TensorFlow~\cite{abadi2016tensorflow} and leverage automatic differentiation, GPU acceleration, distributed training, and many other features (see \cite{sprangers2021probabilistic} for an example). Finally, training GBDT model with neural networks in an end-to-end fashion has recently attracted attention \cite{ivanov2021boost, chen2021convergent} and is worth studying in the context of high-order optimization.





\bibliography{sample}
\newpage
\clearpage
\appendix

\section{Third-Order Split Selection Algorithm}
\begin{algorithm}[h]
    \caption{Selecting optimal split for a tree node}\label{alg:exact-greedy}
    \KwIn{$I$, instances of the node}

    $score\leftarrow 0$\\
    $G^{(i)} \leftarrow \sum_{i\in I} g^{(1)}_{i}$, 
    $G^{(2)} \leftarrow \sum_{i\in I} g^{(2)}_{i}$,
    $G^{(3)} \leftarrow \sum_{i\in I} g^{(3)}_{i}$\\
    \For{$k=1$ {\bfseries to} $m$ }{
       $G^{(i)}_{left}\leftarrow 0$ for $i=1,2,3$\\
       \For{$j $ in sorted($I$, by $\x_{jk}$)}{
            $G^{(i)}_{left}\leftarrow G_{left} + g^{(i)}_j$ for $i=1,2,3$\\
            $G^{(i)}_{right}\leftarrow G^{(i)} - G_{left}^{(i)}$ for $i=1,2,3$\\
            \text{Compute } $\mathcal{L}_{node}$, $\mathcal{L}_{left}$, $\mathcal{L}_{right}$ Eq.~\ref{eq:optimalloss3}\\
            
            $score \leftarrow \max(score, \mathcal{L}_{node} – (\mathcal{L}_{left}+\mathcal{L}_{right}))$\\
       }
    }
    \KwOut{Split with max score}
\end{algorithm}

\section{Full traning curves}\label{app:longertraining}
\begin{figure}[H]
\centering     
\subfigure[Epsilon]{\label{fig:epsilon-10000}\includegraphics[width=.47\textwidth]{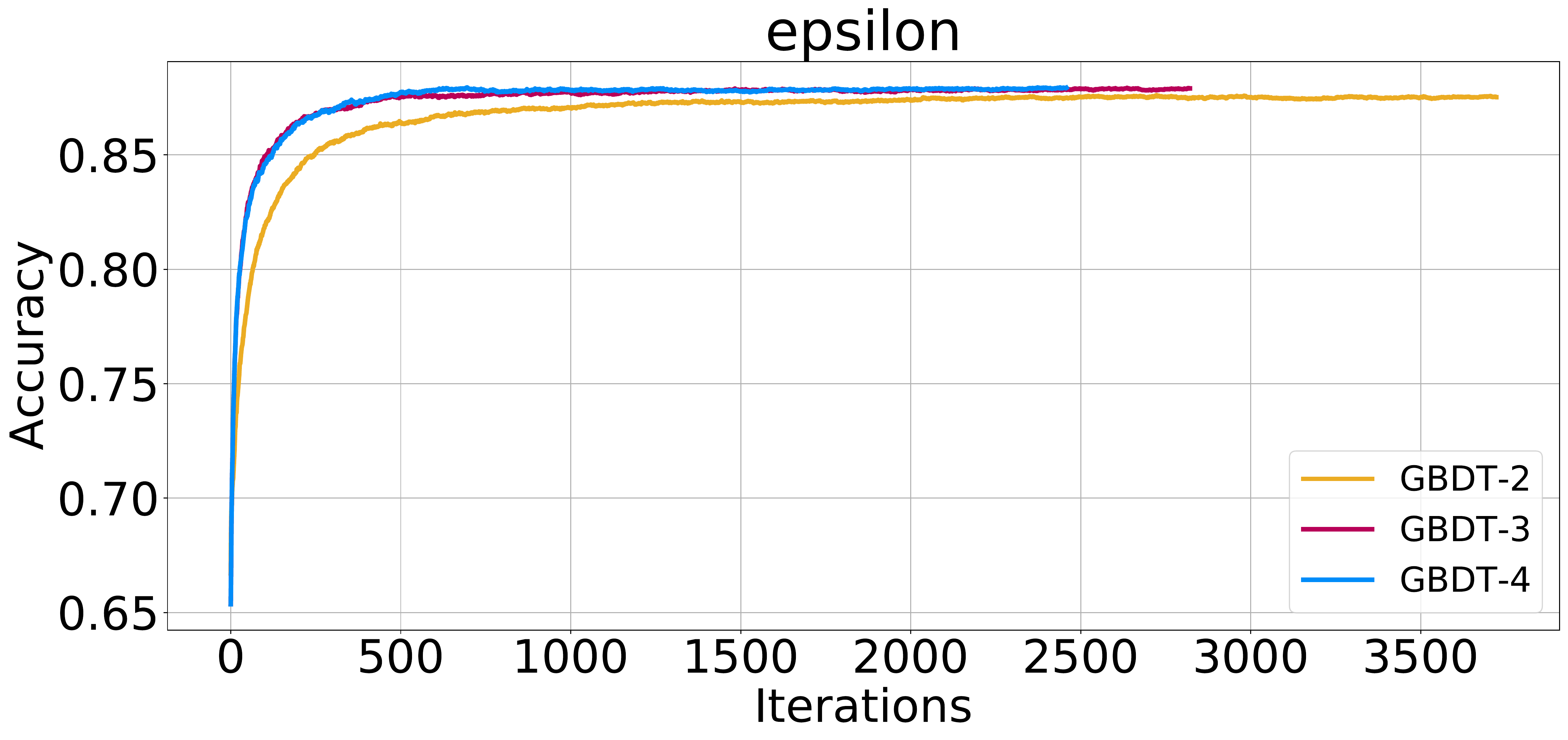}}
\subfigure[Higgs]{\label{fig:Higgs-10000}\includegraphics[width=.47\textwidth]{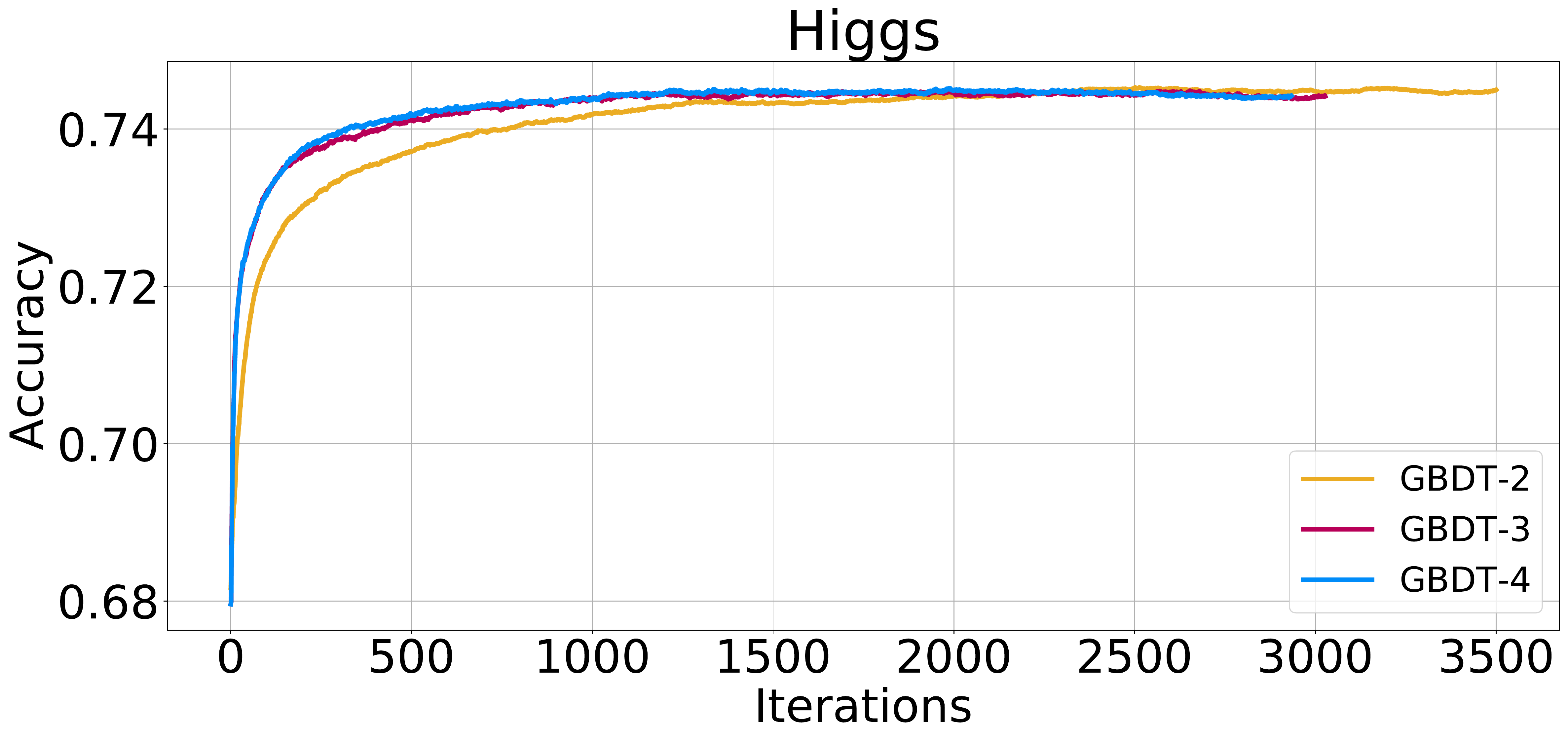}}
\subfigure[MiniBooNE]{\label{fig:MiniBooNE-10000}\includegraphics[width=.47\textwidth]{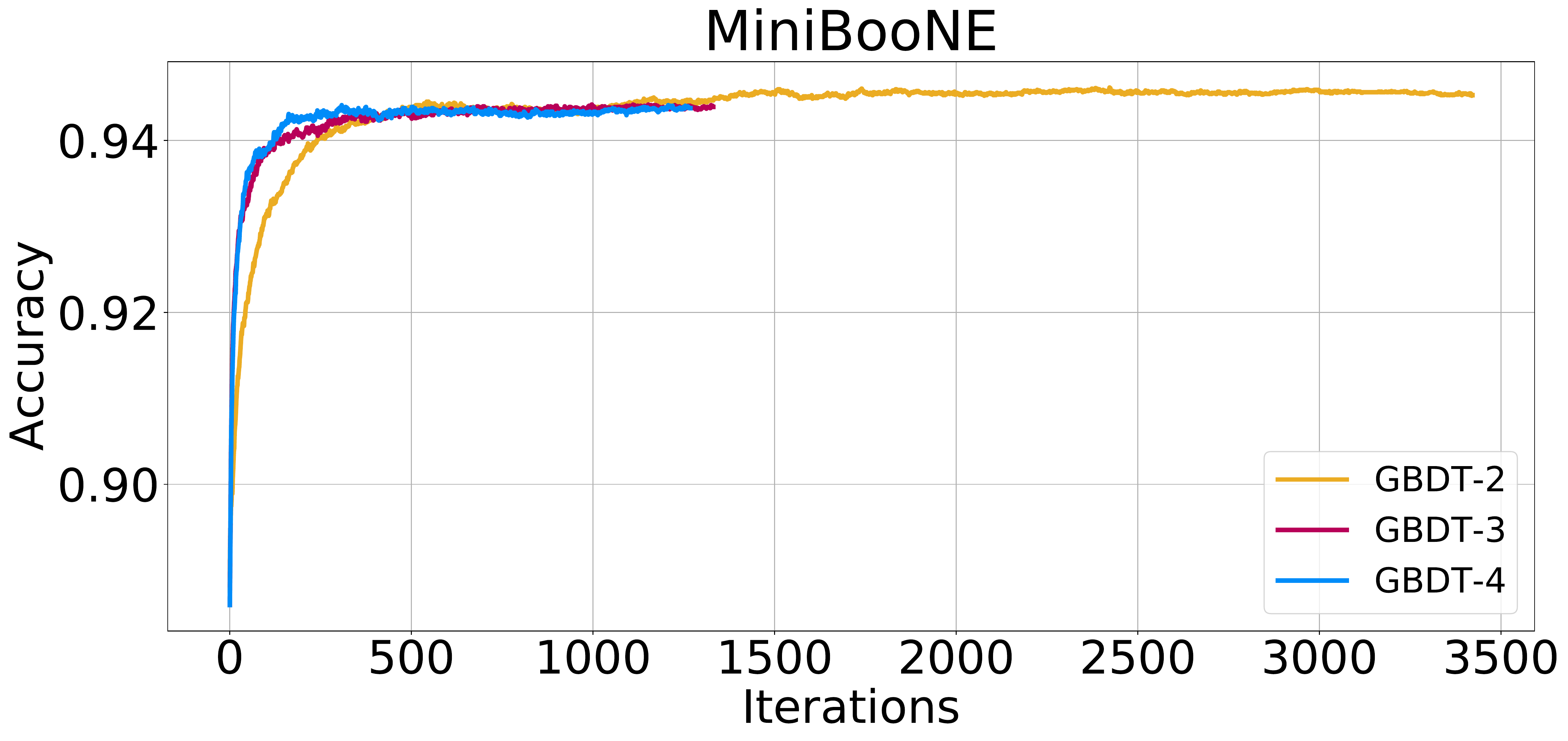}}
\subfigure[covertype]{\label{fig:covertype-10000}\includegraphics[width=.47\textwidth]{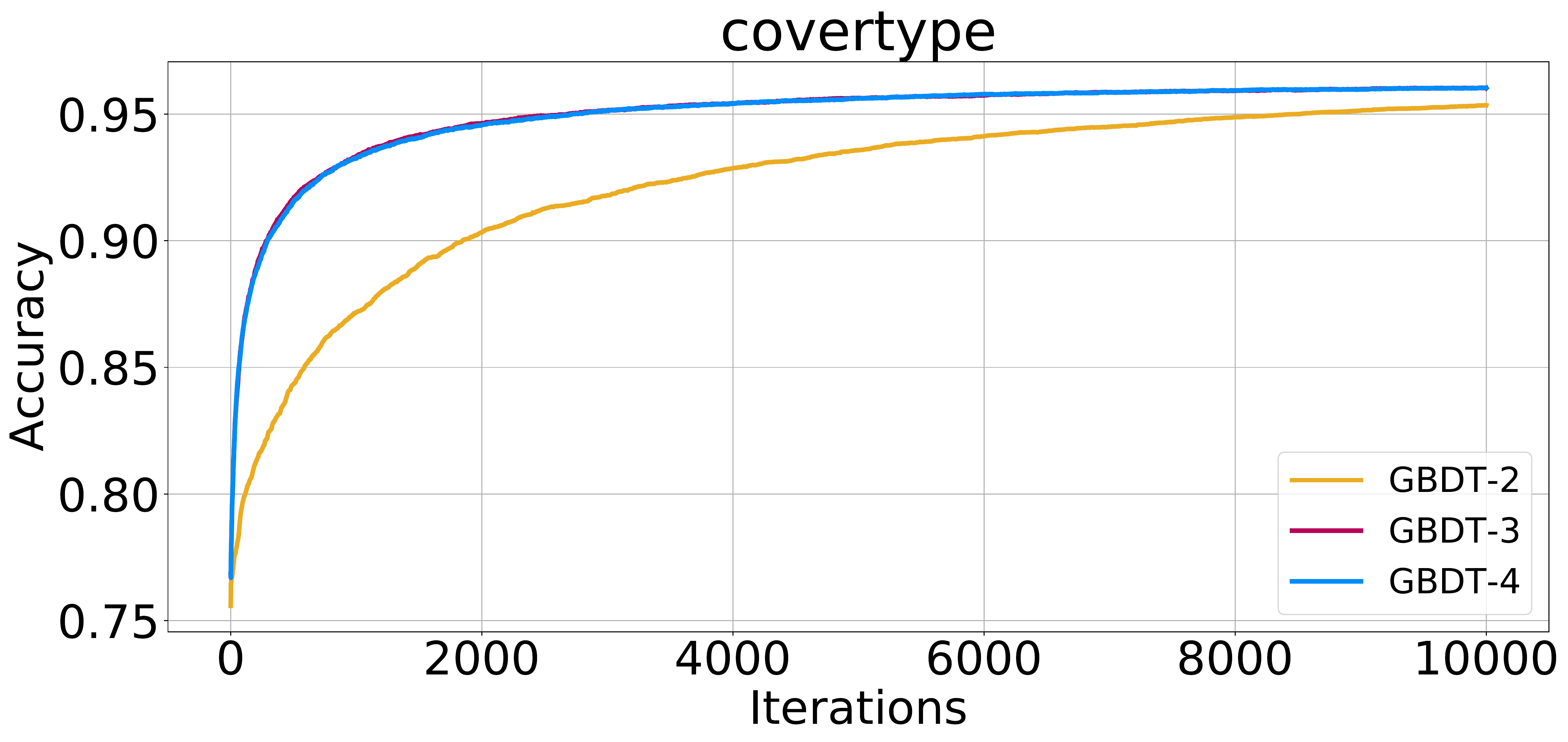}}
\caption{Test accuracy for high-order GBDT models (full training curves).}
\label{fig:appaccuracy}
\end{figure}

\end{document}